\documentclass{article}
\usepackage{spconf,amsmath,graphicx}
\usepackage{subcaption}

\title{Lipreading using Temporal Convolutional Networks}
\name{Brais Martinez\textsuperscript{1}, Pingchuan Ma\textsuperscript{2}, Stavros Petridis\textsuperscript{1,2}, Maja Pantic\textsuperscript{1,2}}
\address{\textsuperscript{1} Samsung AI Research Center, Cambridge, UK\\
\textsuperscript{2} Computing Department, Imperial College London, UK}
\begin{document}
%
\maketitle
\begin{abstract}
Lip-reading has attracted a lot of research attention lately thanks to advances in deep learning. The current state-of-the-art model for recognition of isolated words in-the-wild consists of a residual network and  Bidirectional Gated Recurrent Unit (BGRU) layers. In this work, we address the limitations of this model and we propose changes which further improve its performance. Firstly, the BGRU layers are replaced with Temporal Convolutional Networks (TCN). Secondly, we greatly simplify the training procedure, which allows us to train the model in one single stage. Thirdly, we show that the current state-of-the-art methodology produces models that do not generalize well to variations on the sequence length, and we addresses this issue by proposing a variable-length augmentation. We present results on the largest publicly-available datasets for isolated word recognition in English and Mandarin, LRW and LRW1000, respectively. Our proposed model results in an absolute improvement of 1.2\% and 3.2\%, respectively, in these datasets which is the new state-of-the-art performance.
\end{abstract}

\section{Introduction}
\label{sec:intro}

Visual speech recognition, also known as lip-reading, relies on the lip movements in order to recognise speech without relying on the audio stream. This is particularly useful in noisy environments where the audio signal is corrupted, and can be used in combination with acoustic speech recognisers in order to compensate for the degraded performance due to noise. 

Traditionally, a two-stage approach was followed where features were first extracted from the mouth region, with the Discrete Cosine Transform being the most popular feature extractor, and then fed to a Hidden Markov Model (HMMs) for modeling of the temporal dynamics~\cite{Potamianos2003,Dupont2000,zhou14}. The same two-step approach was also followed in the first deep learning works, where the feature extraction step was replaced by deep autoencoders and HMMs were replaced by Long-Short Term Memory (LSTM) networks \cite{noda2015audio,petridis16,Li2016}. Recently, several end-to-end works have been presented. Such works use either fully connected \cite{petridis2017deepVisualSpeech,wand16, petridis2017end} or convolutional layers~\cite{stafylakis17, shillingford2018large, afouras2018deep, chung16b} to extract features from the mouth region, and then feed them into a recurrent neural network or to attention \cite{chung16b,petridis2018audio} / self-attention architectures \cite{afouras2018deep}.

The state-of-the-art approach for recognition of isolated words is the one proposed in \cite{stafylakis17} which is further refined in~\cite{petridis18} and has been recently extended to a two-stream model in \cite{weng19}. It consists of a 3D convolutional layer followed by an 18-layer Residual Network (ResNet)~\cite{He_2016_CVPR}, a Bidirectional Gated Recurrent Unit (BGRU) network and a softmax layer. It achieves the state-of-the-art performance on the LRW~\cite{chung16b} and LRW1000~\cite{lrw1000} datasets, which are the largest publicly available datasets for isolated word recognition.

In this work, we improve the performance of our state-of-the-art model~\cite{petridis18}. Firstly, we improve the overall performance to achieve a new state-of-the-art. This is achieved by replacing the BGRU layers with a Temporal Convolutional Network~\cite{BaiTCN2018}, which has been shown to achieve similar or even better performance than recurrent layers. Secondly, we simplify the training procedure, reducing training time from 3 weeks to 1 week GPU-time, and avoid relying on a cumbersome 3-stage sequential training. For this purpose, we adopt a cosine scheduler~\cite{sgdr17} and show that training from scratch in one stage is not only feasible, but in fact can produce state-of-the-art results. Finally, we propose a variable-length augmentation procedure to improve the generalization capabilities of the trained model when applied to sequences of varying length\footnote{Length of all LRW videos is 29 frames.}.

\section{Databases}

For the purposes of this study we use the Lip Reading in the
Wild (LRW) \cite{chung16} and LRW1000 \cite{lrw1000} databases which are the largest publicly available lipreading datasets in English and Mandarin, respectively, in the wild. LRW consists of short segments (1.16 seconds) from BBC programs, mainly news and talk shows. There are more than 1000 speakers and a large variation in head pose and illumination and as a consequence is a challenging dataset. The number of words, 500, is also much higher than existing lipreading databases used for word recognition.

 LRW1000 is also a very challenging  dataset due to its large variations in scale, resolution and background clutter. There are 1000 word classes and a total of 718,018 samples with total duration of approximiately 57 hours.

\section{Background}

In the following we describe the methodology of~\cite{petridis18}, as it constitutes the starting point for this work. In particular, we describe the architectural design, of which we maintain the feature encoding part but change the sequence classification layers, and the training procedure, which we thoroughly revamp.

\textbf{Architecture design:} Given a video sequence, the input to the network is a $B\times T \times H \times W$ tensor, each dimension corresponding to batch, frames, height and width, respectively (input images have a single channel indicating gray level). A standard ResNet18 network is used as the video feature encoder, except that the first convolution is substituted by a 3D convolution with kernel $5 \times 7 \times 7$, as proposed in \cite{stafylakis17}. No temporal downsampling is performed throughout the network, and a global spatial average pooling is applied after the last convolution, resulting in a feature output of dimensions $B \times C \times T$, where $C$ indicates the channel dimensionality (512 in this case). Finally, the sequence of feature vectors is fed into a two-layer bidirectional GRU followed by a dense softmax layer~\cite{petridis18}. 

\textbf{Training procedure:} Training follows the three-stage optimization procedure originally proposed in~\cite{stafylakis17}. Training consists of three models trained sequentially, where the previous model is used as initialization to the subsequent one. Firstly, a model is trained with a variant of the network consisting of substituting the BGRU head with a single-layer Temporal Convolutional Network~\cite{BaiTCN2018}. The second model uses the final architecture, the BGRU parameters are randomly initialized, the feature encoding layers are frozen and only the BGRU layers are trained. The third and final step finetunes the full network together. The rationale of this procedure is that RNNs are hard to train, so the feature encoding layers do not have a strong-enough gradient to train properly. Furthermore, \cite{petridis18} uses a custom \textit{reduce and reload} scheduler, so at the end of an epoch, if validation performance drops, the previous checkpoint is loaded and the learning rate is lowered by 3\%. Despite the strong performance demonstrated, the training procedure requires 3 weeks of GPU time to train a single model.

\section{Proposed methodology}
\label{Methodology}

\subsection{Temporal Convolutional Networks} 
\label{ssec:tcn}

Recently, temporal convolutions have emerged as a promising alternative to LSTMs~\cite{BaiTCN2018}, in some cases showing remarkable success on a number of tasks~\cite{wavenet}. A temporal convolution takes a time-indexed sequence of feature vectors as input, and maps it into another such sequence (i.e., the length of the sequence is not altered) through the use of a 1D temporal convolution. Drawing a parallel to the ResNet's basic block, a temporal convolutional block consists of two sets of temporal conv-batchnorm-activation layers, and dropout can be used after each activation. A skip connection/downsample layer is also used, going from the block input to its output. Several such temporal convolutional blocks can be stack sequentially to act as a deep feature sequence encoder. Then, a dense layer is applied to each time-indexed feature vector. Finally, since the aim is sequence classification, a consensus function, in our case a simple averaging, is used.

\begin{figure*}
    \centering
    \begin{subfigure}[b]{0.45\textwidth}
        \includegraphics[width=\textwidth, height=0.6\textwidth, trim=0 1cm 0 0, clip]{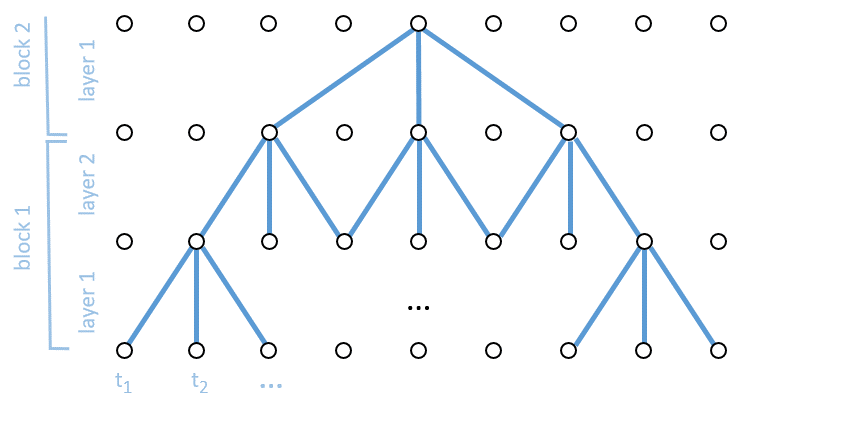}
        \caption{TCN}
        \label{fig:TCN}
    \end{subfigure}
    ~ 
    \begin{subfigure}[b]{0.34\textwidth}
        \includegraphics[width=0.9\textwidth, height=0.8\textwidth]{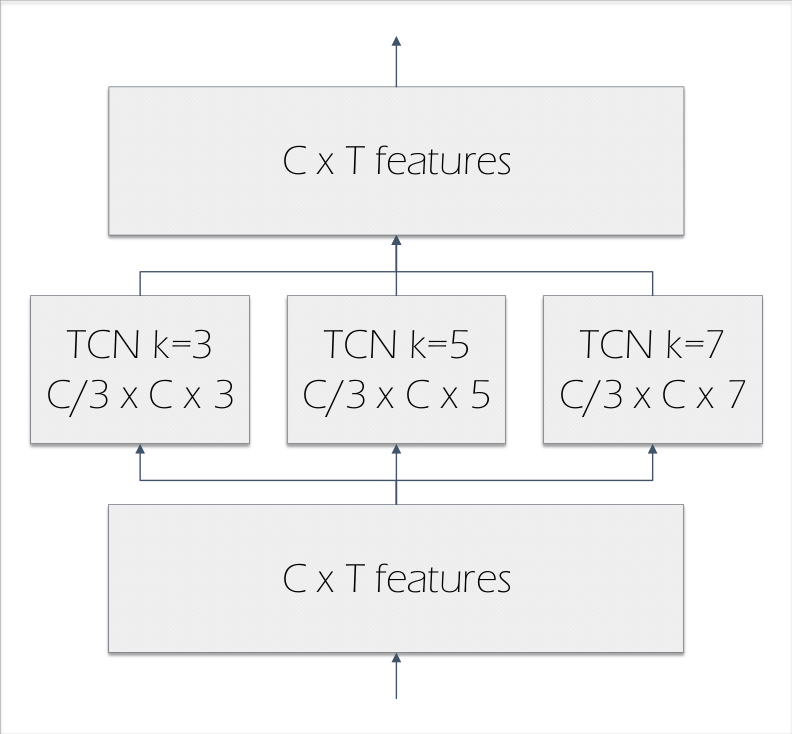}
        \caption{Multiscale TCN}
        \label{fig:MultiTCN}
    \end{subfigure} 
    ~ 
    \begin{subfigure}[b]{0.18\textwidth}
        \includegraphics[ height=1.53\textwidth]{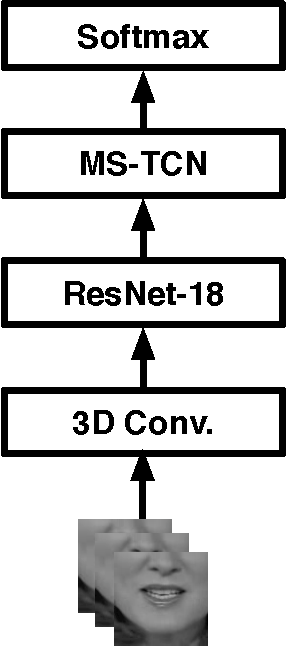}
        \caption{Lip-reading Model}
        \label{fig:LipreadingModel}
    \end{subfigure}
    \caption{(a) Temporal Convolutional Network (TCN). (b) Our Multi-scale Temporal Convolution, which is used in the lip-reading model (c).}\label{fig:modelFig}
\end{figure*}

Dilated convolutions are typically used within TCN to increase the receptive field at a faster rate. In particular, within block $i$, we use a stride of $2^{i-1}$. This architecture is illustrated in Fig.~\ref{fig:TCN}. It is important to note that TCN can be designed to be causal, so at time $t$ only information prior to it is used, or non-causal. Since we are classifying the whole sequence at once, we use the latter design.

Since the input and output of a temporal convolution have the same length, the receptive field of a TCN is defined by the kernel sizes and the stride. Thus, on a standard TCN, all activations at a specific layer share the same temporal receptive field. We would like to provide the network with visibility into multiple temporal scales, in a way that short term and long term information can be mixed up during the feature encoding. To this end, we propose a multi-scale TCN. In this TCN variant, each temporal convolution consists now of several branches, each with different kernel size. When using $n$ branches, each branch has $C/n$ kernels, and their outputs are simply combined through concatenation. In this way, every convolution layer mixes information at several temporal scales. We depict this architecture in Fig.~\ref{fig:MultiTCN}. The full lip-reading model is shown in Fig. \ref{fig:LipreadingModel}.

\subsection{Do lipreading models require complex training?}

Given that we use a purely convolutional architecture, it is reasonable to test whether is possible to train a lipreading model from scratch. We empirically found that in fact it is possible to successfully train the full-fledged model from scratch and achieve state of the art performance. To this end, we adopt a cosine scheduler, which has been shown to be particularly effective~\cite{sgdr17}. Such training leads to competitive performance in 1 week GPU-time.

However, we observe that it is also possible to first pre-train on a subset of the $10\%$ hardest words, which amounts to 50 classes for LRW\footnote{The list of ``hardest words'' is obtained from \cite{stafylakis17}.}. Such initialization allows for faster training, and even yields a small performance improvement. Thus, we adopt this pre-training strategy as it adds a minimal training overhead.

\subsection{Variable length augmentation} 
\label{ssec:varLenAugm}

Models trained on LRW tend to overfit to the dataset scenario, where input sequences are composed of 29 frames and the target word is always at its center. A model trained in such biased setting can memorize these biases, and becomes sensitive to tiny changes being applied to the input. For example, simply removing one random frame from an input sequence results in a significant drop in performance. We propose to avoid this dataset bias by performing variable-length augmentation, by which each input training sequence is cropped temporally at a random point prior and after the target word boundaries. While this change does not lead to a direct improvement on the benchmark at hand, we argue it produces more robust models. We offer some experimental validation of this fact in Section ~\ref{ssec:variable_lengthResults}.

\section{Experimental results}
\label{sec:experiments}

\subsection{Experimental Setting}

\textbf{Pre-processing:} Each video sequence from the LRW dataset is processed by 1) doing face detection and face alignment, 2) aligning each frame to a reference mean face shape 3) cropping a fixed $96\times 96$ pixels wide ROI from the aligned face image so that the mouth region is always roughly centered on the image crop 4) transform the cropped image to gray level, as there does not seem to be a performance difference with respect to using RGB. The mouth ROIs are pre-cropped in the LRW1000 dataset so there is no need for pre-processing.

\looseness-1
We train for 80 epochs using a cosine scheduler and use Adam as the optimizer, with an initial learning rate of $3e-4$, a weight decay of $1e-4$ and batch size of 32. We use random crop of $88\times 88$ pixels and random horizontal flip as data augmentation, both applied in a consistent manner to all frames of the sequence. Finally, we use the standard Cross Entropy loss.

\subsection{Comparison with the current State-of-the-Art}

In this section we compare against the most notable lipreading works in the literature. We attain the state-of-the-art by a wide margin on both LRW by 1.2\%, and LRW1000 by 3.2\% in terms of top-1 accuracy. Furthermore, the current state-of-the-art method on LRW,~\cite{weng19}, relies on much heavier architectures (3D ResNet34), use an ensemble of two networks, and pre-trains on the Kinetics dataset~\cite{kinetics17}, which is extremely compute-intensive. With regards to the baseline method \cite{petridis18}, we achieve 1.9\% better top1 accuracy.
It is also interesting to note that we attain such improvements with a much simpler training strategy.

\begin{table*}
\begin{center}
\begin{tabular}{|l|c|c|c|}
\hline
Method & Backbone & LRW (Accuracy) & LRW-1000 (Accuracy) \\
\hline
LRW \cite{chung16} & VGG-M & 61.1 & -- \\
WAS \cite{chung16b} & VGG-M & 76.2 & --\\
ResNet + LSTM \cite{stafylakis17} & ResNet34$^*$ & 83.0 & 38.2\footnotemark\\
End-to-end AVR \cite{petridis18} & ResNet18$^*$ & 83.4\footnotemark & --\\
Multi-Grained \cite{wang2019} & ResNet34 + DenseNet3D & 83.3 & 36.9 \\
2-stream 3DCNN \cite{weng19} & (3D ResNet34)$\times$ 2 & 84.1 & --\\
\hline
Multi-Scale TCN (Ours) & ResNet18$^*$ & \textbf{85.3} & \textbf{41.4} \\
\hline
\end{tabular}
\vspace{-2mm}
\end{center}
\caption{\it Comparison with state-of-the-art methods in the literature on the LRW and LRW-1000 datasets. Performance is in terms of classification accuracy (the higher the better). We also indicate the backbone employed, as some works either use higher-capacity networks, or use an ensemble of two networks. Networks marked with * use 2D convolutions except for the first being a 3D one.}
\label{tab:sota_visual}
\end{table*}

\footnotetext[3]{Result reported in~\cite{lrw1000}.}
\footnotetext[4]{The paper reports 18\% error, but the author's website shows better results resulting from further model finetuning.}

\subsection{Fixed Length VS Variable Length Training}
\label{ssec:variable_lengthResults}

\begin{table*}[htb]
\begin{center}
\begin{tabular}{|l|c|c|c|c|c|c|}
\hline
Randomly Drop N Frames $\xrightarrow{}$ & N = 0 & N = 1 & N = 2 & N = 3 & N = 4 & N =5 \\
\hline
ResNet18 + BGRU Fixed Length Training & 84.60 & 80.22  & 71.34 & 59.45 & 45.90 & 32.94 \\
ResNet18 + BGRU Variable Length Training & 82.40 & 80.30 & 78.12 & 75.59 & 72.72 & 68.74 \\
ResNet18 + TCN Variable Length Training & 85.30 & 83.54 & 81.18 & 78.70 & 75.68 & 71.49 \\
\hline
\end{tabular}
\vspace{-2mm}
\end{center}
\caption{Classification accuracy of different models on LRW when frames are randomly removed from the test sequences. The model in the first row is the same as the one in \cite{petridis18}. A better accuracy is achieved compared to the one presented in Table \ref{tab:sota_visual} due to the changes in training as explained in section \ref{Methodology}.   }
\label{tab:frame_removal}
\end{table*}

\begin{figure}[ht]
    \begin{center}
    \includegraphics[width=0.9\linewidth, trim=1.3cm 0.5cm 1cm 1.8cm, clip]{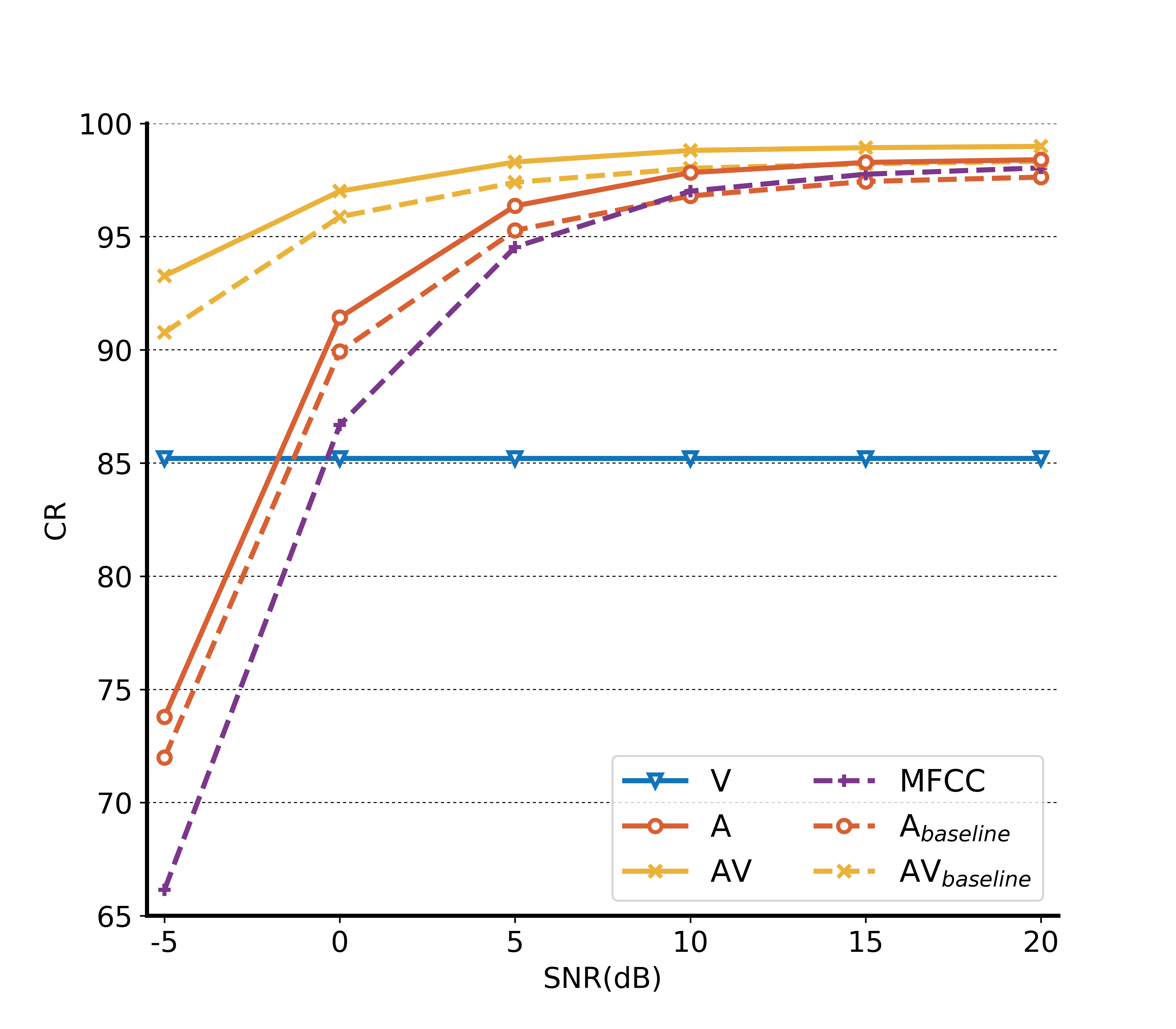}
    \end{center}
    \caption{Performance for audio-only (A), video-only (V) and audio-visual (AV) models under different babble noise levels. The baseline corresponds to the model presented in \cite{petridis18}. }
    \label{fig:av_results}
\end{figure}

The LRW dataset was constructed with several dataset biases that can be exploited to improve on-dataset performance, such as the fact that sequences always have 29 frames, and that words are centered within the sequence. However, it is unrealistic to assume such biases in any real-world test scenario. In order to highlight this issue, we have designed an experiment in which we test the model performance when random frames are removed from the sequence, with the number of frames removed $N$ going from 0 (full model performance) to 5. We further tested the models trained with the variable length augmentation proposed in Sec.~\ref{ssec:varLenAugm}, which is aimed to improve robustness of the model. Results for fixed and variable length training are shown in Table \ref{tab:frame_removal}. All models are trained and evaluated on the LRW dataset. It is clear that the BGRU model trained on fixed length sequences is significantly affected even when one frame is removed during testing. As expected the performance degrades as more and more frames are removed. On the other hand, the BGRU model trained with variable length augmentation is much more robust to frame removals. In the extreme case where 5 frames are removed it results in an absolute improvement of 35.80\% over the fixed length model. However, this robustness is achieved at the expense of reduced performance (84.60\% vs 82.40\%) when the full sequence is fed to the model. On the other hand, the combination of the TCN model with variable length augmentation achieves the same robustness as the BGRU model to frame removals and at the same time leads to superior performance when the full test sequence is presented.

\subsection{Audio-visual experiments}

While lip-reading can be used in isolation, the most useful scenario is when combined with audio to improve performance on noisy environments. In this section we show the performance of our model when trained on audio only, visual only and audio-visual data under varying levels of babble noise. The audio-only and audio-visual models are based on ~\cite{petridis18} but we apply the proposed changes as shown in Fig.~\ref{fig:modelFig}. The performance under different Signal to Noise Ratio (SNR) levels is shown in Fig.~\ref{fig:av_results}. We also compute the performance of a TCN network trained with MFCC features. We use 13 coefficients (and their deltas) using a 40ms window and a 10ms step. Performance of MFCCs is similar to the audio-only model at high SNRs but becomes worse at low SNRs similarly to \cite{petridis18}.

 The audio-visual model is slightly better than the audio-only model at low SNRs but yielding a  clear advantage at  higher levels of noise (low SNR levels). In particular, when using a clean audio signal, the audio-only model attains 1.54\% error rate, while the audio-visual model attains 1.04\%. In the presence of heavy noise, e.g. 0 dB, the audio-visual error rate is 2.92\%, while performance for the audio-only model goes down to 8.57\%. Similarly at -5dB, the audio-visual model achieves an error rate of 6.53\%  significantly outperforming the audio-only model which has an error rate of 26.21\%. We further compare the performance of the audio-only and audio-visual models with respect to our baseline~\cite{petridis18}, showing a clear gain throughout the different noise levels.

\section{Conclusions}

In this work we have presented an improved version of the state-of-the-art model on isolated word recognition. We address the issue of the model not being able to generalise on sequences of varying length by using variable length augmentation. We also replace the BGRUs with temporal convolutional networks which enhance the model's performance. Finally, we simplify the training process so the model can be trained much quicker. The proposed model achieves the new state-of-the-art performance on the LRW and LRW1000 datasets.

\bibliographystyle{IEEEbib}
\bibliography{lipreading}

\begin{thebibliography}{10}

\bibitem{Potamianos2003}
G.~Potamianos, C.~Neti, G.~Gravier, A.~Garg, and A.~W. Senior,
\newblock ``Recent advances in the automatic recognition of audiovisual
  speech,''
\newblock {\em Proceedings of the IEEE}, vol. 91, no. 9, pp. 1306--1326, Sept
  2003.

\bibitem{Dupont2000}
S.~Dupont and J.~Luettin,
\newblock ``Audio-visual speech modeling for continuous speech recognition,''
\newblock {\em IEEE Trans. on Multimedia}, vol. 2, no. 3, pp. 141--151, Sep
  2000.

\bibitem{zhou14}
Z.~Zhou, G.~Zhao, X.~Hong, and M.~Pietik{\"{a}}inen,
\newblock ``A review of recent advances in visual speech decoding,''
\newblock {\em Image and Vision Computing}, vol. 32, no. 9, pp. 590--605, 2014.

\bibitem{noda2015audio}
K.~Noda, Y.~Yamaguchi, K.~Nakadai, H.~G. Okuno, and T.~Ogata,
\newblock ``Audio-visual speech recognition using deep learning,''
\newblock {\em Applied Intelligence}, vol. 42, no. 4, pp. 722--737, 2015.

\bibitem{petridis16}
S.~Petridis and M.~Pantic,
\newblock ``Deep complementary bottleneck features for visual speech
  recognition,''
\newblock in {\em IEEE Int'l Conference on Acoustics, Speech and Signal
  Processing}, 2016.

\bibitem{Li2016}
Y.~Li, Y.~Takashima, T.~Takiguchi, and Y.~Ariki,
\newblock ``Lip reading using a dynamic feature of lip images and convolutional
  neural networks,''
\newblock in {\em IEEE/ACIS Intl. Conf. on Computer and Information Science},
  2016, pp. 1--6.

\bibitem{petridis2017deepVisualSpeech}
S.~Petridis, Z.~Li, and M.~Pantic,
\newblock ``End-to-end visual speech recognition with {LSTMs},''
\newblock in {\em IEEE ICASSP}, 2017, pp. 2592--2596.

\bibitem{wand16}
M.~Wand, J.~Koutnik, and J.~Schmidhuber,
\newblock ``Lipreading with long short-term memory,''
\newblock in {\em IEEE Int'l Conference on Acoustics, Speech and Signal
  Processing}, 2016.

\bibitem{petridis2017end}
S.~Petridis, Y.~Wang, Z.~Li, and M.~Pantic,
\newblock ``End-to-end multi-view lipreading,''
\newblock in {\em BMVC}, 2017.

\bibitem{stafylakis17}
T.~Stafylakis and G.~Tzimiropoulos,
\newblock ``Combining residual networks with {LSTM}s for lipreading,''
\newblock in {\em Interspeech}, 2017.

\bibitem{shillingford2018large}
B.~Shillingford, Y.~Assael, M.~W. Hoffman, T.~Paine, C.~Hughes, U.~Prabhu,
  H.~Liao, H.~Sak, K.~Rao, L.~Bennett, et~al.,
\newblock ``Large-scale visual speech recognition,''
\newblock {\em Interspeech}, 2019.

\bibitem{afouras2018deep}
T.~Afouras, J.~S. Chung, A.~Senior, O.~Vinyals, and A.~Zisserman,
\newblock ``Deep audio-visual speech recognition,''
\newblock {\em IEEE Transactions of Pattern Analysis and Machine Intelligence},
  2018.

\bibitem{chung16b}
J.~S. Chung, A.~Senior, O.~Vinyals, and A.~Zisserman,
\newblock ``Lip reading sentences in the wild,''
\newblock {\em Computer Vision and Pattern Recognition}, 2016.

\bibitem{petridis2018audio}
S.~Petridis, T.~Stafylakis, P.~Ma, G.~Tzimiropoulos, and M.~Pantic,
\newblock ``Audio-visual speech recognition with a hybrid ctc/attention
  architecture,''
\newblock in {\em IEEE Spoken Language Technology Workshop}, 2018, pp.
  513--520.

\bibitem{petridis18}
S.~Petridis, T.~Stafylakis, P.~Ma, F.~Cai, G.~Tzimiropoulos, and M.~Pantic,
\newblock ``End-to-end audiovisual speech recognition,''
\newblock in {\em IEEE Int'l Conference on Acoustics, Speech and Signal
  Processing}, 2018.

\bibitem{weng19}
X.~Weng and K.~Kitani,
\newblock ``Learning spatio-temporal features with two-stream deep {3D CNNs}
  for lipreading,''
\newblock in {\em British Machine Vision Conference}, 2019.

\bibitem{He_2016_CVPR}
Kaiming He, Xiangyu Zhang, Shaoqing Ren, and Jian Sun,
\newblock ``Deep residual learning for image recognition,''
\newblock in {\em Computer Vision and Pattern Recognition}, 2016.

\bibitem{lrw1000}
S.~Yang, Y.~Zhang, D.~Feng, M.~Yang, C.~Wang, J.~Xiao, K.~Long, S.~Shan, and
  X.~Chen,
\newblock ``{LRW-1000:} {A} naturally-distributed large-scale benchmark for lip
  reading in the wild,''
\newblock in {\em Int'l Conf. on Automatic Face {\&} Gesture Recognition},
  2019.

\bibitem{BaiTCN2018}
S.~Bai, J.~Z. Kolter, and V.~Koltun,
\newblock ``An empirical evaluation of generic convolutional and recurrent
  networks for sequence modeling,''
\newblock {\em arXiv:1803.01271}, 2018.

\bibitem{sgdr17}
I.~Loshchilov and F.~Hutter,
\newblock ``{SGDR}: Stochastic gradient descent with warm restarts,''
\newblock in {\em Int'l Conference on Learning Representations}, 2017.

\bibitem{chung16}
J.~S. C. and A.~Zisserman,
\newblock ``Lip reading in the wild,''
\newblock in {\em Asian Conf. on Computer Vision}, 2016.

\bibitem{wavenet}
A.~V.~D. Oord, S.~Dieleman, H.~Zen, K.~Simonyan, O.~Vinyals, A.~Graves,
  N.~Kalchbrenner, A.~Senior, and K.~Kavukcuoglu,
\newblock ``{WaveNet}: {A} generative model for raw audio,''
\newblock {\em arXiv:1609.03499}, 2016.

\bibitem{kinetics17}
J.~Carreira and A.~Zisserman,
\newblock ``Quo vadis, action recognition? a new model and the {Kinetics}
  dataset,''
\newblock {\em Computer Vision and Pattern Recognition}, 2017.

\bibitem{wang2019}
C.~Wang,
\newblock ``Multi-grained spatio-temporal modeling for lip-reading,''
\newblock {\em British Machine Vision Conference}, 2019.

\end{thebibliography}

\end{document}